\documentclass[letterpaper,11pt]{article}
\usepackage{alex}
\usepackage{times}
\usepackage{graphicx}
\usepackage{subfigure}
\usepackage{algorithm}
\usepackage{algorithmic}
\usepackage{hyperref}
\usepackage{amsfonts}
\usepackage{amsmath}
\usepackage{amssymb}
\usepackage{mdwlist}
\renewcommand{\paragraph}[1]{%
  \smallskip
  \noindent{\bfseries #1}}
\title{\bf Local Optimality of User Choices and\\ Collaborative Competitive Filtering}

\author{
Shuang Hong Yang\\
College of Computing\\
Georgia Institute of Technology\\
Atlanta, GA 30332 \\
\texttt{shy@gatech.edu} \\
}

\date{}
\begin{document}

\maketitle
\begin{abstract}
While a user's preference is directly reflected in the interactive
choice process between her and the recommender, this wealth of
information was not fully exploited for learning recommender models.
In particular, existing collaborative filtering (CF) approaches take
into account only the binary events of user actions but totally
disregard the contexts in which users' decisions are made. In this
paper, we propose \emph{Collaborative Competitive Filtering} (CCF),
a framework for learning user preferences by modeling the choice
process in recommender systems. CCF employs a multiplicative latent
factor model to characterize the dyadic utility function. But unlike
CF, CCF models the user behavior of choices by encoding a local
competition effect. In this way, CCF allows us to leverage dyadic
data that was previously lumped together with missing data in
existing CF models. We present two formulations and an efficient
large scale optimization algorithm. Experiments on three real-world
recommendation data sets demonstrate that CCF significantly
outperforms standard CF approaches in both offline and online
evaluations.
\end{abstract}

\section{Introduction}
\label{sec:intro}
Recommender systems have become a core component for today's
personalized online businesses \cite{MurSar03,FleHos07}. With the
abilities of connecting various items (e.g., retailing products,
movies, News articles, advertisements, experts) to potentially
interested users, recommender systems enable online webshops (e.g.
Amazon, Netflix, Yahoo!) to expand the marketing efforts
 from historically a
few best-sellings toward a large variety of long-tail (niche)
products \cite{BryHuSmi03,FleHos07,TanNet10}. Such abilities are
endowed by a personalization algorithm for identifying the
preference of each individual user, which is at the heart of a
recommender system.

Predicting user preference is challenging. Usually, the user and
item spaces are very large yet the observations are extremely
sparse. Learning from such rare, noisy and largely missing evidences
has a high risk of overfitting. Indeed, this \emph{data sparseness}
issue has been widely recognized as a critical challenge for
constructing effective recommender systems.

A straightforward way for building recommender would be to learn a
user's preference based on the prior interactions between her and
the recommender system. Typically, such interaction is an
``opportunity give-and-take" process (\emph{c.f.}
Table~\ref{tab:example}), where at each interaction:
\begin{itemize*}
  \item[1)] a user $u$ inquires the system (e.g. visits a movie recommendation
web site);
  \item[2)] the system offers a set of (personalized)
opportunities (i.e.\ items) $\mathcal{O}=\{i_1,\ldots,i_l\}$ (e.g.
recommends a list of movies of potential interest to the user);
  \item[3)] the user chooses one item $i^*\in\mathcal{O}$
(or more) from these offers and takes actions accordingly (e.g.
click a link, rent a movie, view a News article, purchase a
product).
\end{itemize*}
Somewhat surprisingly, this interaction process has not been
fully-exploited for learning recommenders. Instead, research on
recommender systems has focused almost exclusively on recovering
user preference by completing the matrix of user actions $(u, i^*)$
while the actual contexts in which user decisions are made are
totally disregarded. In particular, Collaborative Filtering (CF)
approaches only captures the action dyads $(u, i^*)$ while the
contextual dyads (i.e.\ $\{(u,i)\}$ for all $i\in\mathcal{O}$ and
$i\ne i^*$) are typically treated as missing data. For example, the
rating-oriented models aim to approximating the ratings that users
assigned to items
\cite{SarwarWWW01,McLaughlinSIGIR04,SalMni08,AgaChe09,ChenKDD09,KorenIEEE09};
the recently proposed ranking-oriented algorithms
\cite{WeiKarLe07,LiuYan08} attempt to recover the ordinal ranking
information derived from the ratings. Although this formulation of
the recommendation problem has led to numerous algorithms which
excel at a number of data sets, including the prize-winning work of
\cite{KorenIEEE09}, we argue here that the formulation is inherently
flawed
--- a preference for \emph{Die Hard} given a generic set of movies
only tells us that the user appreciates action movies; however, a
preference for \emph{Die Hard} over \emph{Terminator} or
\emph{Rocky} suggests that the user might favor Bruce Willis over
other action heroes. In other words, the context of user choice is
vital when estimating user preferences.

\begin{table}[t]
  \begin{minipage}[c]{0.48\columnwidth}
    \centering
    \leftline{\scriptsize
      \begin{tabular}{c|c|c}
        \hline User   & Offer set               & Choice \\
        \hline $u_1$  & [$i_1,i_2,i_3,i_5$]     & $i_2$ \\
        \hline $u_2$  & [$i_2,i_3,i_4,i_5$]     & $i_2$ \\
        \hline $u_3$  & [$i_1,i_3,i_5,i_6$]     & $i_5$ \\
        \hline $u_4$  & [$i_2,i_3,i_4,i_6$]     & $i_3$ \\
        \hline $u_5$  & [$i_1,i_3,i_4,i_5$]     & $i_4$ \\
        \hline $u_6$  & [$i_1,i_4,i_5,i_6$]     & $i_6$ \\
        \hline
      \end{tabular}}
  \end{minipage}
  \hspace{0.02\columnwidth}
  \begin{minipage}[c]{0.48\columnwidth}
    \centering
    \leftline{\scriptsize
      \begin{tabular}{@{\extracolsep{\fill}}c|cccccc}
        \hline &$i_1$  &$i_2$  &$i_3$  &$i_4$  &$i_5$  &$i_6$\\
        \hline
        $u_1$  &$\cdot$       &1      &$\cdot$       &       &$\cdot$       & \\
        $u_2$  &       &1      &$\cdot$       &$\cdot$       &$\cdot$       & \\
        $u_3$  &$\cdot$       &       & $\cdot$      &       &1      &$\cdot$ \\
        $u_4$  &       &$\cdot$       &1      &$\cdot$       &       &$\cdot$ \\
        $u_5$  &$\cdot$       &       &$\cdot$       &1      &$\cdot$       & \\
        $u_6$  &$\cdot$       &       &       &$\cdot$      &$\cdot$       &1 \\
        \hline
      \end{tabular}}
  \end{minipage}

  \caption{An example of user-recommender interactions and
    the derived observation matrix: entries with value
    ``1" denote the action dyads; dyads that are observed
    without user actions (e.g.\ offered by the recommender
    but not picked by the user) are marked with dots
    (``$\cdot$"). CF trains only on the 1 entries while the $\cdot$
    entries are treated as missing data. CCF distinguishes between
    unseen entries and entries marked with dots.
    \label{tab:example}}\vspace{-15pt}
\end{table}

When it comes to modeling of user-recommender interactions, an
important question arises: what is the fundamental mechanism
underlying the user choice behaviors? As reflected by its name,
collaborative filtering is based on the notion of ``collaboration
effects" that \emph{similar items get similar responses from similar
users}. This assumption is essential because by encoding the
``collaboration" among users or among items or both, CF greatly
alleviates the issue of data sparseness and in turn makes more
reliable predictions based on the somewhat \emph{pooled} evidences
across different items/users.

It has long been recognized in psychology and economics that,
besides the effect of collaboration \cite{Byr71,McpLovCoo01},
another mechanism governs users' behavior --- \emph{competition}
\cite{Luc59,McF73,BakBry00}. In particular, items turn to compete
with each other for the attention of users; therefore,
axiomatically, user $u$ will pick the best item $i^*$ (i.e.\ the one
with highest utility) when confronted by the set of alternatives
$\mathcal{O}$. For example, consider a user with a penchant for
action movies by Arnold Schwarzenegger. Given the choice between
\emph{Sleepless in Seattle} and \emph{Die Hard} he will likely
choose the latter. However, when afforded the choice between the
oeuvres of Schwarzenegger, Diesel or Willis, he's clearly more
likely to choose Schwar-zenegger over the works of Willis. To
capture user's preference more accurately, it is therefore
\emph{essential} for a recommender model to take into account such
local competition effect. Unfortunately, this effect is absent in a
large number of collaborative filtering approaches.

In this paper, we present Competitive Collaborative Filtering (CCF)
for learning recommender models by modeling users' choice behavior
in their interactions with the recommender system. Similar to matrix
factorization approaches for CF, we employ a multiplicative latent
factor model to characterize the dyadic utility function (i.e.\ the
utility of an item to a user). In this way, CCF encodes the
collaboration effect among users and items similar to CF. But
instead of learning only the action dyads (i.e.\ $(u,i^*)$ or the
``1" entries in Table~\ref{tab:example}), CCF bases the
factorization learning on the whole user-recommender interaction
traces. It therefore leverages not only the action dyads ($u,i^*$)
but also the dyads in the context without user actions (i.e.\
$(u,i)$ for all $i\in\mathcal{O}$ and $i\ne i^*$ or the dot entries
in Table~\ref{tab:example}), which were treated as potentially
missing data in CF approaches.

To leverage the entire interaction trace for latent factor learning,
we devise probabilistic models or optimization objectives to encode
the local competition effect underlying the user choice process. We
present two formulations with different flavors. The first
formulation is derived from the \emph{multinomial logit model} that
has been widely used for modeling user choice behavior (e.g.\ choice
of brands) in psychology \cite{Luc59}, economics \cite{Man75,McF73}
and marketing science \cite{GuaLit08}. The second formulation
relates closely to the ordinal regression models in content
filtering \cite{HerGraObe99} (e.g.\ web search ranking).
Essentially, both formulations attempt to encodes ``local optimality
of user choices" to encourage that every opportunity $i^*$ taken by
a user $u$ be locally the best in the context of the opportunities
$\mathcal{O}$ offered to her. From a machine learning viewpoint, CCF
is a hybrid of \emph{local} and \emph{global} learning, where a
global matrix factorization model is learned by optimizing a local
context-aware loss function. We discuss the implementation of CCF,
establish efficient learning algorithms and deliver an package that
allows distributed optimization on streaming data.

Experiments were conducted on three real-world recommendation data
sets. First, on two dyadic data sets, we show that CCF improves over
standard CF models by up to 50+\% in terms of offline top-$k$
ranking. Furthermore, on a commercial recommender system, we show
that CCF significantly outperform CF models in both offline and
online evaluations. In particular, CCF achieves up to 7\%
improvement in offline top-$k$ ranking and up to 13\% in terms of
online click rate prediction.

\paragraph{Outline:}
\S\ref{sec:backgroud} describes the problem formulation, the
backgrounds and motivates CCF. \S\ref{sec:ccf} presents the detailed
CCF models, learning algorithms and our distributed implementation.
\S\ref{sec:experiment} reports experiments and results.
\S\ref{sec:relatedwork} reviews related work and
\S\ref{sec:conclusion} summarizes the results.

\section{Preliminaries}
\label{sec:backgroud}

\subsection{Problem formulation}
\label{sec:def} Consider the user-system interaction in a
recommender system: we have users $u \in \mathcal{U}:=\{1, 2,
\ldots, U\}$ and items $i \in \mathcal{I}=\{1,2,\ldots, I\}$; when a
user $u$ visits the site, the system recommends a set of items
$\mathcal{O} = \{i_1,\ldots, i_l\}$ and $u$ in turn chooses a
(possibly empty) subset $\mathcal{D}\subseteq \mathcal{O}$ from
$\mathcal{O}$ and takes actions accordingly (e.g.\ buys some of the
recommended products). For ease of explanation, let us temporarily
assume $\mathcal{D} = \cbr{i^*}$, i.e.\ $\Dcal$ is not empty and
contains exactly one item $i^*$. More general scenarios shall be
discussed later.

To build the recommender system, we record a collection of
historical  interactions in the form of
$\{(u_t,\mathcal{O}_t,\mathcal{D}_t)\}$, where $t$ is the index of a
particular interaction session. Our goal is to generate
recommendations $\mathcal{O}_{\tilde{t}}$ for an incoming visit
$\tilde{t}$ of user $u_{\tilde{t}}$ such that the user's
satisfaction is maximized. Hereafter, we refer to $\mathcal{U}$ as
\textit{user space}, $\mathcal{I}$ as \textit{item space},
$\mathcal{O}_t$ as \textit{offer set} or \emph{context},
$\mathcal{D}_t$ as \textit{decision set}, and $i^*$ as a
\emph{decision}.

A key component of a recommender system is a model $r(u,i)$ that
characterizes the utility of an item $i\in\mathcal{I}$ to a user
$u\in\mathcal{U}$, upon which recommendations for a new inquiry from
user $u$ could be done by simply ranking items based on $r(u,i)$ and
recommending the top-ranked ones. Collaborative filtering is by far
the most well-known method for modeling such dyadic responses.

\subsection{Collaborative filtering}
\label{sec:cf} In collaborative filtering we are given observations
of dyadic responses $\{(u,i,y_{ui})\}$ with each $y_{ui}$ being an
observed response (e.g.\ user's rating to an item, or indication of
whether user $u$ took an action on item $i$). The whole mapping:
\begin{align*}
  (u,i) \rightarrow y_{ui}\text{ where } u\in \mathcal{U}, i\in
  \mathcal{I}
\end{align*}
constitutes a large matrix $Y \in
\mathcal{Y}^{\vert\mathcal{U}\vert\times \vert\mathcal{I}\vert}$.
While we might have millions of users and items, only a tiny
proportion (considerably less than 1\% in realistic datasets) of
entries are observable. Note the subtle difference in terms of the
data representation: while we record entire sessions, CF only
records the dyadic responses.

Collaborative filtering explores the notion of ``collaboration
effects", i.e., similar users have similar preference to similar
items. By encoding collaboration, CF pools the sparse observations
in such a way that for predicting $r(u,i)$ it also borrows
observations from other users/items. Generally speaking, existing CF
methods fall into either of the following two categories.

\paragraph{Neighborhood models.}
A popular class of approaches to CF is based on propagating the
observations of responses among items or users that are considered
as neighbors. The model first defines a similarity measure between
items / users. Then, an unseen response between user $u$ and item
$i$ is approximated based on the responses of neighboring users or
items \cite{SarwarWWW01,McLaughlinSIGIR04}, for example, by simply
averaging the neighboring responses with similarities as weights.

\paragraph{Latent factor models.}
This class of methods learn predictive latent factors to estimate
the missing dyadic responses. The basic idea is to associate latent
factors\footnote{Throughout this paper, we assume each latent factor
$\phi$ contains a constant component so as to absorb
user/item-specific offset into latent factors.}, $\phi_u
\in\mathbb{R}^k$ for each user $u$ and $\phi_i\in\mathbb{R}^k$ for
each item $i$, and assume a multiplicative model for the dyadic
response,
\begin{align}
  p(y_{ui}|u,i) = p(y_{ui} | r_{ui}; \Theta),\notag
\end{align}
where $\Theta$ denotes the set of hyper-parameters, the utility is
assumed as a multiplicative function of the latent factors,
\[r(u,i)=\phi_u^\top \phi_i.\]
This way the factors could explain past responses and in turn make
prediction for future ones. This model implicitly encodes the
Aldous-Hoover theorem \cite{Kallenberg05} for exchangeable matrices
-- $y_{ui}$ are independent of each other given $\phi_u$ and
$\phi_i$. In essence, it amounts to a low-rank approximation of the
matrix $Y$ that naturally embeds both users and items into a vector
space in which the distances directly reflect the semantic
relatedness.

To design a concrete model
\cite{AiroldiNIPS08,MillerNIPS09,SinghECML08}, one needs to specify
a distribution for the dependence. Afterwards, the model boils down
to an optimization problem. For example two commonly-used
formulations are:
\begin{description*}
\item[- $\ell_2$ regression] The most popular learning formulation is to
  minimize the $\ell_2$ loss within an empirical risk minimization
  framework \cite{KorenKDD08,SalMni08}:
  \begin{align*}
\min_{\phi} \sum_{(u,i)\in \Omega} (y_{ui} - \phi_u^\top\phi_i)^2 +
\lambda_{\mathcal{U}}\sum_{u\in\mathcal{U}}\vert\vert\phi_u\vert\vert^2
+
\lambda_{\mathcal{I}}\sum_{i\in\mathcal{I}}\vert\vert\phi_u\vert\vert^2,
\end{align*}
  where $\Omega$ denotes the set of ($u,i$) dyads for which the
  responses $y_{ui}$ are observed, $\lambda_{\mathcal{U}}$ and
  $\lambda_{\mathcal{I}}$ are regularization weights.
\item[- Logistic] Another popular formulation
  \cite{MillerNIPS09,AgaChe09} is to use logistic regression by
  optimizing the cross-entropy
  \begin{align*}
    \min_{\phi} & \sum_{(u,i)\in \Omega}
    \log \sbr{1 + \exp(-\phi_u^\top \phi_i)}\\
    &+ \lambda_{\mathcal{U}}\sum_{u\in\mathcal{U}}\vert\vert\phi_u\vert\vert^2
    + \lambda_{\mathcal{I}}\sum_{i\in\mathcal{I}}\vert\vert\phi_u\vert\vert^2
  \end{align*}
\end{description*}

\subsection{Motivating discussions}
Collaborative filtering approaches have made substantial progresses
and are currently the state-of-the-art techniques for recommender
system.  However, we argue here that CF approaches might be a bit
lacking in several aspects. First of all, although data sparseness
is a big issue, CF does not fully leverage the wealth of user
behavior data. Take the user-recommender interaction process
described in \S\ref{sec:def} as an example (c.f.
Table~\ref{tab:example}), CF methods typically use only the action
dyad $(u,i^*)$ of each session while other dyads $\{(u,i)\vert i\in
\mathcal{O}, i\ne i^*\}$ are treated missing and totally
disregarded, which could be wasteful of the invaluable learning
resource because these non-action dyads are not totally useless, as
shown by the experiments in this paper.

Secondly, most existing CF approaches learn user preference
collaboratively by either approximating the dyadic responses
$\{y_{ui^*}\}$
\cite{SarwarWWW01,McLaughlinSIGIR04,SalMni08,AgaChe09,ChenKDD09,KorenIEEE09}
or preserving the ordinal ranking information derived from the
dyadic responses \cite{WeiKarLe07,LiuYan08}; none of them models the
user choice behavior in recommender systems. Particularly, as users
choose from competing alternatives, there is naturally a local
competition effect among items being offered in a session. Our work
show that this effect could be an important clue for learning user
preference.

Because latent factor models are very flexible and could be
under-determined (or over-parameterized) even for rather moderate
number of users/items. With the above two limitations, CF approaches
are vulnerable to over-fitting \cite{AgaChe09,KorenIEEE09}.
Particularly, while most existing CF models might learn consistently
on user ratings (numerical value typically with five levels) if
given enough training data, they usually perform poorly on binary
responses. For example, for the aforementioned interaction process
(\emph{c.f.} Table~\ref{tab:example}), the response $y_{ui}$ is
typically a binary event indicating whether or not item $i$ was
accepted by the user $u$. With the non-action dyads being ignored,
the responses are exclusively positive observations (either
$y_{ui}=1$ or missing). As a result, we will obtain an
overly-optimistic estimator that biases toward positive responses
and predicts positive for almost all the incoming dyads (See
\S\ref{sec:dyad_exp} for empirical evidences).
%Indeed, lack of sufficient variant in user responses is one of the
%principal reason why most existing research on recommendation has
%been focused on rating data rather than user behavior data.

\section{Collaborative competitive\\ filtering}
\label{sec:ccf}
We present a novel framework for recommender learning by modeling
the system-user interaction process. The key insight is that the
contexts $\mathcal{O}_t$ in which user's decisions are made should
be taken into account when learning recommender models. In practice,
a user $u$ could make different decisions when facing different
contexts $\mathcal{O}_t$. For instance, an item $i$ would not have
been chosen by $u$ if it were not presented to her at the first
place; likewise, user $u$ could choose another item if the context
$\mathcal{O}_t$ changes such that a better offer (e.g., a more
interesting item) is presented to her.

In this section, we describe the framework of
collaborative-competitive filtering. We start with some axiomatic
views of the user choice behaviors. Following that, we present the
learning formulation of CCF. We then develop the optimization
algorithms and implementation techniques. We close the section with
a discussion of useful extensions.

\subsection{Local optimality of user choices}
Formally, the individual choice process (i.e.\ user-recommender
interactions) in a recommender system can be viewed as an instance
of the \emph{opportunity give-and-take} (GAT) process.

\smallskip\noindent{\textsc{\textsf{\textbf{Definition} [GAT]:}}} \textit{An
opportunity give-and-take process is a process of interactions among
an agent $u$, a system $S$ and a set of opportunities $\mathcal{I}$;
at an interaction $t$:}\vspace{-6pt}
\textit{\begin{itemize*}
  \item[-]  $u$ is given a set of opportunities $O_t\subset\mathcal{I}$ by $S$;
  \item[-]  $u$ makes the decision by takeing one of the opportunities:
  $i^*_t\in\mathcal{O}_t$;
  \item[-]  Each opportunity $i\in\mathcal{O}_t$ could potentially give $u$ a
  revenue (utility) of $r_{ui}$ if being taken or 0 otherwise.
\end{itemize*}}
Note that we assume the agent is {\it a priori} not aware of all the
items, and only through the recommender $S$ can she get to know the
items, therefore other items that are not in $\mathcal{O}_t$ is
unaccessible to $u$ at interaction $t$. This is reasonable
considering that the number of item is usually very large. Moreover,
we assume an agent $u$ is a rational decision maker: she knows that
her choice of item $i$ will be at the expense of others
$i^\prime\in\mathcal{O}_t$, therefore she compares among
alternatives before making her choice. In other words, for each
decision, $u$ considers both \textsf{revenue} and
\textsf{opportunity cost}, and decides which opportunity to take
based on the potential \textsf{profit} of each opportunity in
$\mathcal{O}$. Specifically, the opportunity cost $c_{ui}$ is the
potential loss of $u$ from taking an opportunity $i$ that excludes
her to take other opportunities: $c_{ui} = \max\{r_{ui^\prime}:
i^\prime\in\mathcal{O}\setminus i\}$; the profit $\pi_{ui} = r_{ui}
- c_{ui}$ is the net gain of an decision. By drawing the rational
decision theory \cite{Luc59}, we present the following principle of
individual choice behavior.

\smallskip\noindent{\textsc{\textsf{\textbf{Proposition}:}}} \textit{A
rational decision is a decision maximizing the profit: $i^*$ =
$\arg\max_{i\in\mathcal{O}} \pi_{ui}$.}\smallskip

This proposition implies the constraint of ``local optimality of
user choice", a local competitive effect restricting that the agent
$u$ always chooses the offer that is locally optimal in the context
of the offer set $\mathcal{O}_t$.

\subsection{Collaborative competitive filtering}
The local-optimality principle induces a constraint which could be
translated to an objective function for recommender learning:
\begin{align}\label{eq:local}
&\forall i^*\in\mathcal{D}_t,~~~ r_{ui^*} \geqslant
\max\{r_{ui}\vert i\in \mathcal{O}_t\setminus\mathcal{D}_t\}\notag\\
\text{or}~~ &P( i^*\text{ is taken}) = P( r_{ui^*} \geqslant
\max\{r_{ui}\vert i\in \mathcal{O}_t\setminus\mathcal{D}_t\}).
\end{align}

This objective is, however, problematic. First, the inequality
constraint restricts the utility function only up to an arbitrary
order-preserving transformation (e.g.\ a monotonically increasing
function), and hence cannot yield a unique solution (e.g.\ point
estimation) \cite{Man75}. Second, optimization based on the induced
objective is computationally intractable due to the $\max$ operator.
To this end, we present two surrogate objectives, which both are
computationally efficient and show close connections to existing
models.

\subsubsection{Softmax model}
Our first formulation is based on the random utility theory
\cite{Luc59,Man75} which has been extensively used for modeling
choice behavior in economics \cite{McF73} and marketing science
\cite{GuaLit08}. In particular, we assume the utility function
consists of two components $r_{ui} + e_{ui}$, where: (1) $r_{ui}$ is
a deterministic function characterizing the intrinsic interest of
user $u$ to item $i$, for which we use the latent factor model to
quantify $r_{ui}=\phi_u^\top\phi_i$; (2) the second part $e_{ui}$ is
a stochastic error term reflecting the uncertainty and complexness
of the choice process\footnote{The error term essentially accounts
for all the subtle, uncertain and unmeasurable factors that
influence user choice behaviors, for example, a user's mood, past
experience, or other factors (e.g., whether the decision is made in
a hurry, together with her friends, or totally unconsciously)}.
Furthermore, we assume the error term $e_{ui}$ is an independently
and identically distributed Weibull (extreme point) variable:
\begin{align*}
\Pr(e_{ui}\leqslant\epsilon) = e^{-e^{-\epsilon}}.
\end{align*}

Together with the local-optimality principle, these two constraints
yield the following \emph{multinomial logit model}
\cite{McF73,Man75,GuaLit08}:
\begin{align}\label{eq:logit}
 p(i^*=i\vert u, \mathcal{O})
  =\frac{e^{r_{ui}}}{\sum_{j\in\mathcal{O}} e^{r_{uj}}}
  \text{ for all } i\in\mathcal{O}.
\end{align}
Intuitively, this model enforces the local-optimality constraint by
using the \emph{softmax} function as a surrogate of \emph{max}.

Given a collection of training interactions $\{(u_t,\mathcal{O}_t,
i^*_t)\}$, the latent factors can be estimated using penalized
maximum likelihood via
\begin{align}
  \label{eq:softmax}
  \min_{\phi} ~ &
  \sum_{t} \log \Bigl[\sum_{i\in\mathcal{O}_t}\exp(\phi_{u_t}^\top\phi_i)\Bigr] -
  \phi_{u_t}^\top \phi_{i^*_t} \\
  \nonumber
  &+ \lambda_{\mathcal{U}}\sum_{u\in\mathcal{U}}\vert\vert\phi_u\vert\vert^2
  + \lambda_{\mathcal{I}}\sum_{i\in\mathcal{I}}\vert\vert\phi_u\vert\vert^2.
\end{align}
While the above formulation is a convex optimization w.r.t. $r_{ui}$
as each of the objective terms in Eq.(\ref{eq:softmax}) is strongly
concave, it is nonconvex w.r.t. the latent factors $\phi$. We
postpone the discussion of optimization algorithms to
\S\ref{sec:op}.

\subsubsection{Hinge model}
Our second formulation is based on a simple reduction of the
local-optimality constraint. Note that, from Eq(\ref{eq:local}), it
follows that:
\begin{align}
P(i=i^*\vert u,
\mathcal{O})&=P((r_{ui^*}-r_{ui})>(e_{ui}-e_{ui^*}),~\forall
i\in\mathcal{O})\notag\\
&\leqslant
P((r_{ui^*}-\bar{r}_{u\tilde{i}})>(\bar{e}_{u\tilde{i}}-e_{ui^*})),\notag
\end{align}
where
$\bar{r}_{u\tilde{i}}=\frac{1}{\vert\mathcal{O}\vert-1}\sum_{i\in\mathcal{O}\setminus
i^*}r_{ui}$ is the average potential utility that $u$ could possibly
gain from the non-chosen items. Intuitively, the above model
encourages that the utility difference between choice and non-chosen
items, $r_{ui^*} - \bar{r}_{u\tilde{i}}$, to be nontrivially greater
than random errors. Based on this notion, we present the following
formulation which views the task as a pairwise preference learning
problem \cite{HerGraObe99} and uses the non-choices averagely as
negative preferences.
\begin{align}
  \label{eq:hinge}
  \min_{\theta, \xi} ~ & \sum_t \xi_{t} +
  \lambda_{\mathcal{U}}\sum_{u\in\mathcal{U}}\vert\vert\phi_u\vert\vert^2
  +
  \lambda_{\mathcal{I}}\sum_{i\in\mathcal{I}}\vert\vert\phi_u\vert\vert^2\\
  \nonumber
  \text{s.t.:~} &
  r_{ui^*_t} - \frac{1}{|\mathcal{O}_t| - 1} \sum_{i \in
    \mathcal{O}_t\backslash\cbr{i^*_t}} r_{ui} \geq 1 - \xi_t
  \text{ and } \xi_t \geq 0.
\end{align}
This formulation is directly related to the maximum score estimation
\cite{Man75} of the multinomial logit model Eq(\ref{eq:logit}).
Intuitively, it directly reflects the insight that user decisions
are usually made by comparing alternatives and considering the
\emph{difference} of potential utilities. In other words, it learns
latent factors by maximizing the marginal utility between user
choice and the average of non-choices.

Again, the optimization is convex w.r.t. $r_{ui}$, but nonconvex
w.r.t. the latent factors, therefore the standard optimization tools
such as the large variety of RankSVM \cite{HerGraObe99} solvers are
not directly applicable.

\subsubsection{Complexity}
It is worth noting that our CCF formulations have an appealing
linear complexity,
$O(\vert\mathcal{I}\vert\times\vert\mathcal{O}\vert)$, where the
offer size $\vert \mathcal{O}\vert$ is typically a very small
number. For example, Netflix recommends $\vert \mathcal{O}\vert=7$
movies for each visit, and Yahoo!\ frontpage highlights $\vert
\mathcal{O}\vert=4$ hot news for each browser. Therefore, CCF has
the same-order complexity as the rating-oriented CF models. Note
that the ranking-oriented CF approaches \cite{WeiKarLe07,LiuYan08}
are much more expensive -- for each user $u$, the learning
complexity is quadratic $O(\vert\mathcal{I}\vert^2)$ as they learn
preference of each user by comparing every pair of the items.
\subsection{Learning algorithms}
\label{sec:op}
As we have already mentioned, due to the use of bilinear terms, both
of the two CCF variants are nonconvex optimization problems
regardless of the choice of the loss functions. While there
\emph{are} convex reformulations for some settings they tend to be
computationally inefficient for large scale problems as they occur
in industry --- the convex formulations require the manipulation of
a full matrix which is impractical for anything beyond thousands of
users.

Moreover, the interactions between user and items change over time
and it is desirable to have algorithms which process this
information incrementally. This calls for learning algorithms that
are sufficiently efficient and preferably capable to update
dynamically so as to reflect upcoming data streams, therefore
excluding offline learning algorithms such as classical SVD-based
factorization algorithms \cite{KorenIEEE09} or spectral eigenvalue
decomposition methods \cite{LiuYan08} that involve large-scale
matrices.

We use a distributed stochastic gradient variant with averaging
based on the Hadoop MapReduce framework. The basic idea is to
decompose the objectives in \eq{eq:softmax} or \eq{eq:hinge} by
running stochastic optimization on sub-blocks of the interaction
traces in parallel in the Map phase, and to combine the results for
$\phi_i$ in the Reduce phase. The basic structure is analogous to
\cite{ChenKDD09,ZinWeiSmo10}.

\paragraph{Stochastic Optimization.}
We derive a stochastic gradient descent algorithm to solve the
optimization described in Eq\eq{eq:softmax} or Eq\eq{eq:hinge}. The
algorithm is computationally efficient and decouplable among
different interactions and users, therefore amenable for parallel
implementation.

The algorithm loops over all the observations and updates the
parameters by moving in the direction defined by negative gradient.
Specifically, we can carry out the following update equations on
each machine separately:
\begin{itemize*}
\item For all $i\in\mathcal{O}_t$ do $\phi_i \leftarrow \phi_i -
  \eta \sbr{l'(\phi_u^\top \phi_i) \phi_u + \lambda_{\mathcal{I}}\phi_i}$.
\item For each $u$ do $\phi_u \leftarrow \phi_u -
  \eta \sbr{\sum_{i \in \mathcal{O}_t} l'(\phi_u^\top \phi_i) \phi_i
    + \lambda_{\mathcal{U}}\phi_u}$.
\end{itemize*}
Here $\eta$ is the learning rate\footnote{We carry out an annealing
procedure to discount $\eta$ by a factor of 0.9 after each
iteration, as suggested by \cite{KorenKDD08}.}. The gradients are
given by:
\begin{align}\label{eq:grad}
l'_{\mathrm{Softmax}}(r_{ui})&=\frac{\exp(r_{ui})}{\sum_{j\in\mathcal{O}}\exp(r_{uj})}
- \delta_{i,i^*} \\
l'_{\mathrm{Hinge}}(r_{ui})&=-\frac{\abr{\mathcal{O}} \delta_{i,i^*}
    -1}{\abr{\mathcal{O}} - 1} H(1 - r_{ui^*} + \bar{r}_{u\tilde{i}})
\end{align}
where $H(\cdot)$ is the Heaviside function, i.e.\ $H(x) = 1$ if $x >
0$ and $H(x) = 0$ otherwise.\footnote{In our implementation, we
approximate this by the continuous function $\frac{1}{1+e^{-100
    x}}$. This helps with convergence.}

\paragraph{Feature Hashing.}
A key challenge in learning CCF models on large-scale data is that
the storage of parameters as well as observable features requires a
large amount of memory and a reverse index to map user IDs to memory
locations. In particular in recommender systems with hundreds of
millions of users the memory requirement would easily exceed what is
available on today's computers (100 million users with 100 latent
feature dimensions each amounts to 40GB of RAM). We address this
problem by implementing feature hashing \cite{WeiKilAni09} on the
space of matrix elements. In particularly, by allowing random
collisions and applying hash mapping to the latent factors (i.e.\
$\phi$), we keep the entire representation in memory, thus greatly
accelerating optimization.

\subsection{Extensions}
\label{sec:extensions}

We now discuss two extensions of CCF to address the fact that in
some cases users choose not to respond to an offer at all and that
moreover we may have observed features in addition to the latent
representation discussed so far.

\subsubsection*{Sessions without response}

In establishing the CCF framework for modeling the user choice
behavioral data, we assumed that for each user-system interaction
$t$, the decision set $\mathcal{D}_t$ contains at least one item.
This assumption is, however, not true in practice. A user's visit at
a recommender system does not always yields an action. For example,
users frequently visit online e-commerce website without making any
purchase, or browse a news portal without clicking on an ad.
Actually, such nonresponded visits may account for a vast majority
of the traffics that an recommender system receives. Moreover,
different users may have different propensities for taking an
action.  Here, we extend the multinomial logit model to modeling
both responded and nonresponded interactions,
$(u_t,\mathcal{O}_t,i^*_t)$ and $(u_t, \mathcal{O}_t, \emptyset)$
respectively.

This is accomplished by adding a scalar $\theta_u$ for each user $u$
to capture the \emph{action threshold} of user $u$. We assume that,
at an interaction $t$, user $u_t$ takes an effective action only if
she feels that the overall quality of the offers $\mathcal{O}_t$ are
good enough and worth the spending of her attention. In keeping with
the logistic model this means that
\begin{align}
  \label{eq:softmax_ext}
  p(i^*=i| u, \mathcal{O})
  =\frac{\exp(\phi_u^\top\phi_{i})}{\exp(\theta_u) + \sum_{j\in\mathcal{O}}\exp(\phi_u^\top\phi_j)}
\end{align}
for all $i \in \mathcal{O}$ and the probability of no response is
given by the remainder, that is by
$\frac{\exp(\theta_u)}{\exp(\theta_u) +
  \sum_{j\in\mathcal{O}}\exp(\phi_u^\top\phi_j)}$.
In essence, this amounts to a model where the `non-response' has a
certain reserve utility that needs to be exceeded for a user to
respond. We may extend the hinge model in the same spirit (we use a
trade-off constant $C > 0$ to calibrate the importance of the
non-responses).
\begin{align}
  \nonumber
  \min_{\phi, \xi, \epsilon, \theta} ~ & \sum_t \xi_{t}+C\sum_t \varepsilon_{t}+
  \lambda_{\mathcal{U}}\sum_{u\in\mathcal{U}}\vert\vert\phi_u\vert\vert^2
  +
  \lambda_{\mathcal{I}}\sum_{i\in\mathcal{I}}\vert\vert\phi_u\vert\vert^2\\
  \nonumber
  \text{subject to } & r_{u_ti^*_t} - \bar{r}_{u_t\tilde{i}} -\theta_u
  \geqslant 1 - \xi_{t} \text{ for all } i^*\in \mathcal{D}_t
  \text{ if } \mathcal{D}_t \neq \emptyset \\
  \nonumber
  &\theta_u -r_{u_ti} \geqslant 1 - \varepsilon_{t},~~\forall i\in
  \mathcal{O}_t\text{ if }\mathcal{D}_t = \emptyset \\
  &r_{ui}=\phi_u^\top\phi_i \text{ and }
  \xi_{t}\geqslant0 \text{ and } \varepsilon_{t}\geqslant0
  \label{eq:hinge_ext}
\end{align}

\begin{table}[h!]
  \caption{Statistics of data sets.
  \label{tab:data}}
  \bigskip
\centering
\begin{tabular}{lcccc}
\hline &{\bf\#user}  & {\bf\#item} &{\bf\#dyads}  &{\bf offer size}\\
\hline  {\sf Social} & 1.2M & 400 & 29M & -\\
        {\sf Netflix-5star} & 0.48M & 18K & 100M & -\\
        {\sf News} & 3.6M & 2.5K & 110M & 4\\
\hline
\end{tabular}
\end{table}

\subsubsection*{Content features}
In previous sections, we use a plain latent factor model for
quantifying utility, i.e.\ $r_{ui}=\phi_u^\top\phi_i$. A known
drawback \cite{AgaChe09} of such model is that it only captures
dyadic data (responses), and therefore generalizes poorly for
completely new entities, i.e.\ unseen users or items, of which the
observations are missing at the training stage. Here, we extend the
model by incorporating content features. In particular, we assume
that, in addition to the latent features $\phi$s, there exist some
observable properties $x_u\in \mathbb{R}^{m}$ (e.g. a user's
self-crafted registration files) for each user $u$, and
$x_i\in\mathbb{R}^n$ (e.g.\ a textual description of an item) for
each item $i$. We then assume the utility
  $r_{ui}$ as a function of both types of features (i.e.\ observable and latent):
\begin{align*}
  r_{ui}\sim p(r_{ui}\vert \phi_u^\top \phi_i+ x_u^\top M x_i; \Theta)
\end{align*}
where the matrix $M\in \mathbb{R}^{m\times n}$ provides a bilinear
form for characterizing the utility based on the content features of
the corresponding dyads. This model integrates both collaborative
filtering \cite{KorenIEEE09} and content filtering \cite{ChuPar09}.
On the one hand, if the user $u$ or item $i$ has no or merely
non-informative observable features, the model degrades to a
factorization-style utility model \cite{SalMni08}. On the other
hand, if we assume that $\phi_u$ and $\phi_i$ are irrelevant, for
instance, if $i$ or $j$ is totally new to the system such that there
is no interaction involving either of them as in a cold-start
setting, this model becomes the classical content-based relevance
model commonly used in, e.g.\ webpage ranking
\cite{ZheZhaZhaChaetal08}, advertisement targeting \cite{ChenKDD09},
and content recommendation \cite{ChuPar09}.

\section{Experiments}
\label{sec:experiment}

We report experimental results on two test-beds. First, we evaluate
the CCF models with CF baselines on two dyadic data sets with
simulated choice contexts. The choice of simulated data generated
from CF datasets was made since we are unaware of any
\emph{publicly} available datasets directly suitable for CCF.
Furthermore, we extend our evaluation to a more strict setting based
on user-system interaction session data from a commercial
recommender system.

\subsection{Dyadic response data}
\label{sec:dyad_exp}

\begin{figure}
  \centering
  \includegraphics[width=0.7\columnwidth]{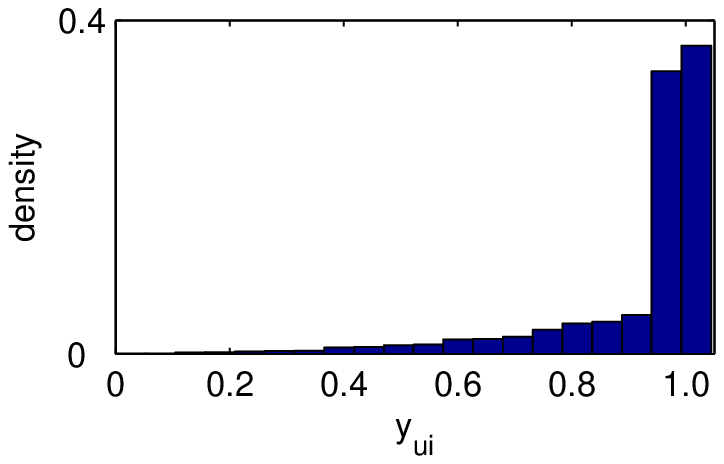}
  \includegraphics[width=0.7\columnwidth]{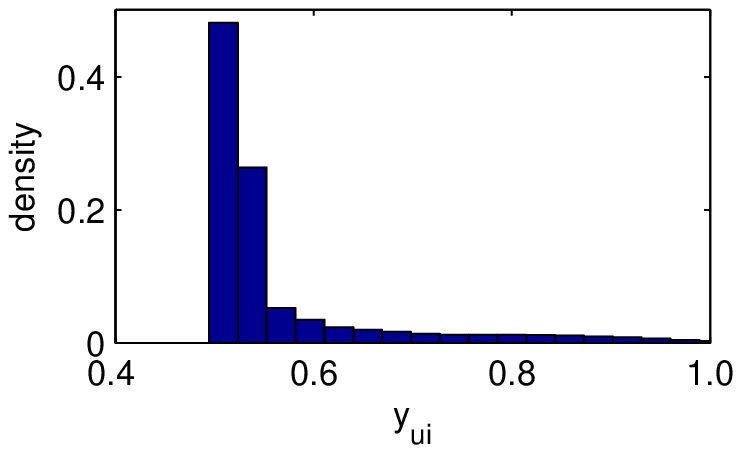}
  \caption{Histograms of the predicted dyadic responses
    $\cbr{\hat{y}_{ui}}$: while the predictions by CF (top) are
    over-optimistically concetrated on positive responses (i.e.\
    predicting ``relevant" for all possible dyads), the results obtained
    by CCF (bottom) demonstrate a more realistic power-law property.
    \label{fig:degree}}
\end{figure}

We use dyadic data with binary responses, i.e.\ $\{(u,i,y_{ui})\}$
where $y_{ui} \in \cbr{1, \mathrm{missing}}$.  We compare different
recommender models in terms of their top-$k$ ranking performance.

\paragraph{Social network data.}
The first data set we used was collected from a commercial social
network site, where a user expresses her preference for an item with
an explicit indication of ``like". We examine data collected for
about one year, involving hundreds of millions of users and a large
collection of applications, such as games, sports, news feeds,
finance, entertainment, travel, shopping, and local information
services. Our evaluation focuses on a random subset consisting of
about 400 items, 1.2 million users and 29 million dyadic responses
(``like" indications).

\paragraph{Netflix 5 star data.}
For the sake of reproducibility of our results, we also report
results on a data set derived from the Netflix prize
data\footnote{\url{http://www.netflixprize.com}}, one of the most
famous public data sets for recommendation. The Netflix data set
contains 480K users and 18K movies. We derive binary responses by
considering only 5-star ratings as ``positive" dyads and treating
all the others as missing entries.

For both data sets, we randomly split the data into three pieces,
one for training, one for testing and the other for validation.

\paragraph{Evaluation metrics.}
We assess the recommendation performance of each model by comparing
the top suggestions of the model to the true actions taken by a user
(i.e.\ ``like" or 5-star). We consider three measures commonly used
for accessing top-$k$ ranking performance in the IR community:
\begin{description*}
  \item[AP] is the \emph{average precision}. AP@$n$
  averages the precision of the top-$n$ ranked list of each
  query (e.g. user).
  \item[AR] or \emph{average recall} is the average
  recall of the top-$n$ rank list of each query.
  \item[nDCG] or \emph{normalized Discounted Cumulative Gain}
  is the normalized position-discounted precision score. It gives larger
  credit to top positions.
\end{description*}
For all the three metrics we use $n = 5$ since most social networks
and movie recommendation sites recommend a similar number of items
for each user visit.

\begin{table}
  \caption{Top-$k$ ranking performance on two binary dyadic data sets with
  simulated contexts.
  \label{tab:simulate}}
  \bigskip
\centering
\begin{tabular}{|ll|c|c|c|}
\hline \multicolumn{2}{|c|}{Model}   &  {AP@5}    & {AR@5}    & {nDCG@5}\\
\hline \multicolumn{5}{|c|}{{\textsf{Social}}}\\
\hline  {CF} & $\ell_2$  & 0.448        & 0.230         & 0.475   \\
        {CF} & Logistic  & 0.449        & 0.230         & 0.476\\
        {CCF} & Softmax  & {\bf 0.688}  & {\bf 0.261}   & {\bf 0.704}   \\
        {CCF} & Hinge    & 0.686        & 0.260         & 0.702   \\
\hline \multicolumn{5}{|c|}{{\textsf{Netflix-5star}}}\\
\hline  {CF}  & $\ell_2$  & 0.135           & 0.022         & 0.145  \\
        {CF}  & Logistic  & 0.135           & 0.023         & 0.146 \\
        {CCF} & Softmax   & {\bf 0.186}     & {\bf 0.033}   & {\bf 0.189}  \\
        {CCF} & Hinge     & 0.185           & 0.032         & 0.188  \\
\hline
\end{tabular}
\end{table}

\paragraph{Evaluation protocol.}
We compare the two CCF models (i.e.\ Softmax and Hinge) with the two
standard CF factorization models (i.e.\ $\ell_2$ and Logistic)
described in \S\ref{sec:cf}. For dyadic data with binary responses,
the Logistic CF model amounts to the state-of-the-art
\cite{MillerNIPS09,AgaChe09}.

We adopt a fairly strict top-$k$ ranking evaluation. For each user,
we assess the top results out of a total preference ordering of the
whole item set. In particular, for each user $u$, we consider all
the items as candidates; we compute the three measures based on the
comparison between the ground truth (the set of items in the test
set that user $u$ actually liked) and the top-5 suggestions
predicted by each model. For statistical consistency, we employ a
cross-validation style procedure. We learn the models on training
data with parameters tuned on validation data, and then apply the
trained models to the test data to assess the performance. All three
measures reported are computed on test data only, and they are
averaged over five random repeats (i.e.\ random splits of the
data).\footnote{ \emph{Note that the contextual information (the
offer set
  $\mathcal{O}_t$ for each interaction $t$) is missing for both of the
  two dyadic data sets.} We choose the datasets anyway to ensure that
the results (at least on the Netflix dataset) can be repeated by
other research groups. Results on interaction data are reported in
\S\ref{sec:ccf-experiments}.}

To render the data compatible with CCF we simulate a fixed-size
pseudo-offer set for each interaction. Specifically, for every
positive observation, e.g.\ $y_{ui} =1$, we randomly sample a
handful set of missing (unobserved) entries
$\{y_{ui^\prime}\}_{i^\prime=1:m}$. These sampled dyads are then
treated as non-choices, and together with the positive dyad, they
are used as the offer set for the current session. In our
experiments, we choose $m=9$ pseudo non-choices; in other words, we
assume the offer size $\vert\mathcal{O}_t\vert$ =10.

\paragraph{Results and analysis.}
We report the mean scores in Table~\ref{tab:simulate}. Since the
dataset are fairly large the standard deviations of all values were
below $0.001$. Consequently we omitted the latter from the results.
As can be seen from the table, CCF dramatically outperforms CF
baselines on both data sets. In terms of AP@5, the two CCF models
gain about {52.8\%--53.6\%} improvements compared to the two CF
models on the \textsf{Social} data, and by {37.0\%--37.8\%} on
\textsf{Netflix-5 star}. Similar comparisons apply to the nDCG@5
measure. And in terms of the AR@5, CCF models outperform CF
competitors by up to 13.5\% on \textsf{Social}, and 30\% on
\textsf{Netflix-5 star} data. All these improvements are
statistically highly significant.
Note that these results are quite consistent: both CF models perform
comparably with each other on both data sets; the performance of the
two CCF variants is also comparable; between the two groups, there
are noticeable gaps.

One argument we made in this paper for motivating our work is that
since the CF models disregard the context information and only
learns on positive (action) dyads, they almost inevitably yield
overly-optimistic predictions (i.e.\ predicting positive for all
possible dyads). We hypothesize that such estimation \emph{bias} is
one of the key reasons for the inability of CF models in learning
binary dyadic data. As an empirical validation, in
Figure~\ref{fig:degree}, we plot the histograms of the predicted
dyadic responses $\hat{y}_{ui}$ (i.e.\ entries of the diffused
matrices) obtained by a CF model ($\ell_2$) and a CCF model
respectively.\footnote{Similar results obtained with other losses.}
As we can see, the CF model indeed predicts ``positive" for most (if
not all) dyads; in contrast, the results obtained by the CCF model
demonstrate a more realistic power-law distribution
\cite{Faloutsos99}.\footnote{Note that the distribution
  starts at around 0.5 instead of 0, which is consistent with our
  intuitions that there is actually \emph{no} truly ``irrelevant" item
  for a user -- any item has potential utility for a user; user choose
  one over another based on the relative preference rather than
  absolute utility. This is true especially in this era of information
  explosion, where a user is typically facing so many alternatives
  that she can only pick the one she likes most while ignoring the
  others.}

In reality, each user can only afford to ``like" a few items out of
a huge amount of alternatives. This power-law property is crucial
for information filtering because we are intended to identify a few
truly relevant items by \emph{filtering} out many many irrelevant
ones. A power-law recommender is desirable in a way analogous to a
filter with narrow-bandwidth, which effectively filters the noises
(i.e.\ irrelevant items) and only let the true signal (i.e.\
relevant items) pass to the endnode (i.e.\ users).

\subsection{User-system interaction data}
\label{sec:ccf-experiments}
We now move on to a more realistic evaluation by applying CCF to
real user-system interaction data. We evaluate CCF in both an
offline test and an online test while comparing its results to both
CF baselines.

\paragraph{Data.} We collected a large-scale set of user-system
interaction traces from a commercial article (News feeds)
recommender system. In each interaction, the system offers four
personalized articles to the visiting user, and the user chooses one
of them by clicking to read that article. The recommendations are
dynamically changing over time even during the user's visit. The
system regularly logs every click event of every user visit. It also
records the articles being presented to users at a series of
discrete time points. To obtain a context set for each user-system
interaction, we therefore trace back to the closest recording time
point right before the user-click, and we use the articles presented
at that time point as the offer set for the current session. We
collected such interaction traces from logged records of over one
month.  We use a random subset containing 3.6 million users, 2500
items and over 110 million interaction traces. Learning an effective
recommender on this data set is particularly challenging as the
article pool is dynamically refreshing, and each article only has a
lifetime of several hours --- it only appears once within a
particular day, is pulled out from the pool afterward never to
appear again.

\paragraph{Evaluation protocol.} We consider the following two evaluation
settings, one offline and the other offline.
\begin{description*}
  \item[Offline evaluation] Similar to the evaluations presented
  in \S\ref{sec:dyad_exp}, we evaluate the learned recommender models in terms
  of the top-$k$ ranking performance on a hold-out test subset. We follow
  the same configurations in \S\ref{sec:dyad_exp} and use
  the three ranking measures, i.e.\ AP@n, AR@n and nDCG@n as the evaluation
  metrics. Note that here we use $n=4$ instead of 5, because it is the
  default recommendation size used in the news recommender system.
\item[Online evaluation] We further conduct an online test. In
  particular.  for each incoming interaction, we use the trained
  models to predict which item among the four recommendations will
  be taken by the user. This prediction is of crucial importance
  because one of the key objectives for a recommender system is to
  maximize the traffic and monetary revenue by lifting the
  click-through rate.
\end{description*}

\begin{table}
  \caption{Offline test (top-$k$ ranking performance) on user-system interaction data.
  \label{tab:offline}}
  \bigskip
\centering
\begin{tabular}{|ll|c|c|c|}
\hline \multicolumn{2}{|c|}{Model}  &{AP@4}    &{AR@4}    &{nDCG@4}\\
\hline \multicolumn{5}{|c|}{{\textsf{30\% Training}}}\\
\hline  {CF}     & $\ell_2$      & 0.245            & 0.261         & 0.255   \\
        {CF}     & Logistic      & 0.246            & 0.263         & 0.257\\
        {CCF}    & Softmax       & {\bf 0.262}      & {\bf 0.278}   & {\bf 0.274}   \\
        {CCF}    & Hinge         & 0.261            & {\bf 0.278}   & 0.273   \\
\hline \multicolumn{5}{|c|}{\textsf{{50\% Training}}}\\
\hline  {CF}     & $\ell_2$      & 0.250            & 0.273         & 0.268  \\
        {CF}     & Logistic      & 0.252            & 0.276         & 0.269  \\
        {CCF}    & Softmax       & {\bf 0.266}      & {\bf 0.285}   & {\bf 0.278} \\
        {CCF}    & Hinge         & 0.265            & {\bf 0.285}   & 0.277 \\
\hline \multicolumn{5}{|c|}{\textsf{{70\% Training}}}\\
\hline  {CF}     & $\ell_2$      & 0.253            & 0.275         & 0.271  \\
        {CF}     & Logistic      & 0.253            & 0.276         & 0.274  \\
        {CCF}    & Softmax       & {\bf 0.267}      & {\bf 0.287}   & {\bf 0.280}  \\
        {CCF}    & Hinge         & {\bf 0.267}      & 0.286         & {\bf 0.280} \\
\hline
\end{tabular}
\end{table}

\paragraph{Offline test results.} In this setting, we train each model on progressive
proportions of 30\%, 50\% and 70\% randomly-sampled training data
respectively, and evaluate each trained model in terms of offline
top-$k$ ranking performance. The results are reported in
Table~\ref{tab:offline}. The two CCF models greatly outperform the
two CF baselines in all the three evaluation metrics. Specifically,
CCF models gain up to 6.9\% improvement over the two CF models in
terms of average precision; up to 6.5\% in terms of average recall,
and up to 7.5\% in terms of nDCG. We also conducted a $t$-test with
a standard $0.05$ significance level. The hypothesis tests indicate
that all the improvements obtained by CCF are significant.

It is worth noting that the improvements obtained by CCF compared to
CF baselines are especially evident when the training data are
sparser (e.g.\ using only 30\% of training data). This observation
empirically validates our argument that the contexts contain
substantial useful information for learning recommender models
especially when the dyadic action responses are scarce.

\begin{table}[h!]
  \caption{Online test (predicted click probability) on user-system interaction data.
  \label{tab:online}}
  \bigskip
\centering
\begin{tabular}{|ll|c|c|c|}
\hline\multicolumn{2}{|c|}{Model}&{30\%train}   &{50\%train}    &{70\%train}\\
\hline\multicolumn{2}{|c|}{Random}&\multicolumn{3}{|c|}{0.250}\\
\hline
        {CF}     & $\ell_2$      & 0.337        & 0.343         & 0.347   \\
        {CF}     & Logistic      & 0.341        & 0.345         & 0.347\\
        {CCF}    & Softmax       & 0.376        & 0.384         & {\bf 0.391}   \\
        {CCF}    & Hinge         &{\bf 0.377}   & {\bf 0.385}   & {\bf 0.391}   \\
\hline
\end{tabular}
\end{table}

\begin{figure}[htbp]
\centering
\includegraphics[width=\columnwidth]{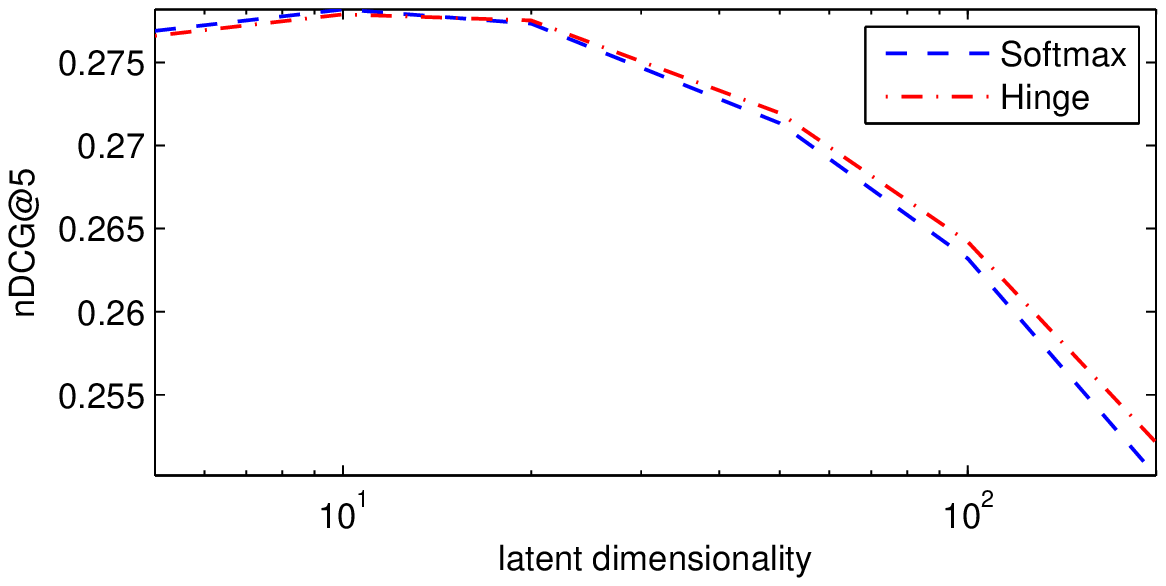}
\includegraphics[width=\columnwidth]{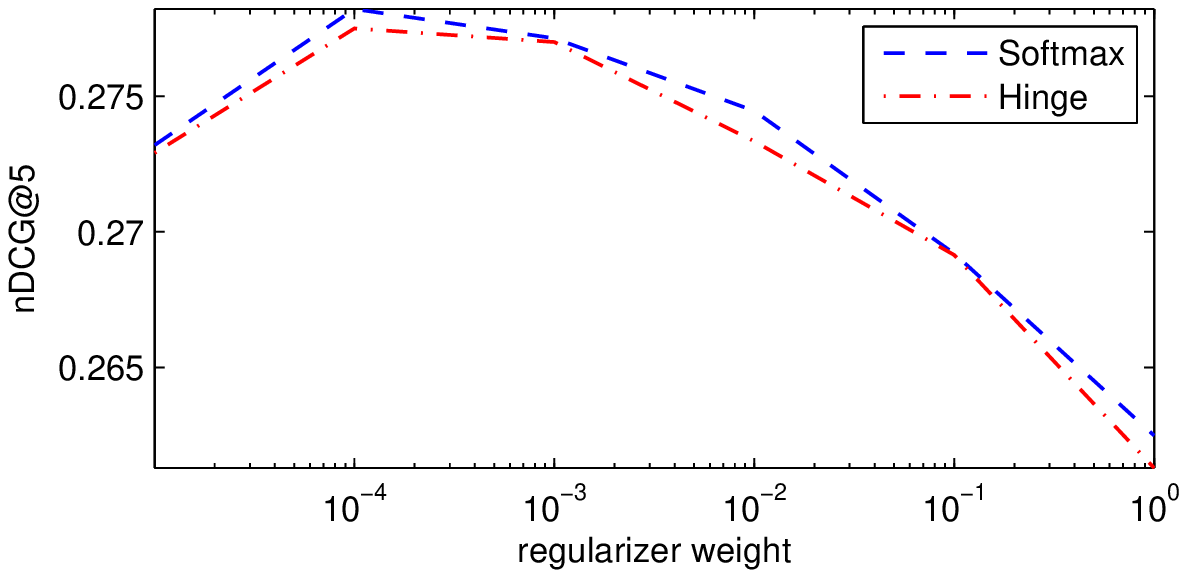}
\caption{Offline top-$k$ ranking performance (nDCG@5) as a function
of latent dimensionality (top) and regularization weight (bottom).
\label{fig:para}}
\end{figure}

The offline results obtained by CCF are quite satisfactory. For
example, the average precision is up to 0.276, which means, out of
the four recommended items, on average 1.1 are truly ``relevant"
(i.e.\ actually being clicked by the user). This performance is
quite promising especially considering that most of the articles in
the content pool are transient and subject to dynamically updating.

\paragraph{Online test results.} We further evaluate the online performance
of each compared model by assessing the predicted click rates.
Click-rate is essential for an online recommender system because it
is closely-related to both the traffic and the revenue of a webshop.
In our evaluation, for each of the incoming visits
$(u_t,\mathcal{O}_t,i^*_t)$, we use the trained models to predict
the user choice, i.e. we ask the question: ``among all the offered
items $i\in\mathcal{O}_t$, which one will most likely be clicked?"
We use the trained model to rank the items in the offer set, and
compare the top-ranked item with the item that was actually taken
(i.e.\ $i_t^*$) by user $u_t$. We evaluate the results in terms of
the prediction accuracy.

The results are given in Table~\ref{tab:online}. Because the size of
each offer set in the current data set is 4, a random predictor
yields 0.25. As seen from the table, while all the four models
obtain significantly better predictions than the random predictor,
the two CCF models further greatly outperform the two CF models.
Specifically, we observe 11.3\%--12.7\% improvements obtained by CCF
models compared to the two CF competitors. These results are quite
significant especially considering the dynamic property of the
system.

\paragraph{Impact of parameters.} The performance of the two CCF
models is affected by the parameter settings of the latent
dimensionality, $k$, as well as the regularization weights,
$\lambda_\mathcal{I}$ and $\lambda_\mathcal{U}$. In
Figure~\ref{fig:para}\footnote{Due to heavy computational
consumptions, these results are obtained on a relatively small
subset of data.}, we illustrate how the offline top-$k$ ranking
performance changes as a function of these parameters, where we use
the same value for both $\lambda_\mathcal{I}$ and
$\lambda_\mathcal{U}$. Here we only reported the results with nDCG@5
measure because the results show similar shapes when other measures
(including the click rate) are used. As can be seen from the Figure,
the nDCG curves are typically in the inverted U-shape with the
optimal values achieved at the middle. In particular, for both the
two CCF models, the dimensionality around 10 and regularization
weight around 0.0001 yield the best performance, which is also the
default parameter setting we used in obtaining our reported results.

%\paragraph{Scalability.}

\paragraph{Nonresponded sessions.} In Section~\ref{sec:extensions} we presented two models for encoding
nonresponded interactions, e.g.\ a user visits the News website but
does not click any of the recommended articles. These approaches are
promising because compared to the responded sessions, the
nonresponded ones are typically much more plentiful and if learned
successfully, this wealth of information has a potential to
alleviating the critical data-sparse issue in recommendation.

Unfortunately, due to the data-logging mechanism of the News
recommender system, we were unable to obtain such nonresponded
interactions (this is subject to future work).  Instead, for a
preliminary test, we conducted evaluation on a small set of
\emph{pseudo} nonresponded sessions that are derived from the
responded ones. In particular, we hold out a randomly-sampled subset
of sessions; for each of these sessions, we hide the item being
clicked by the user, and use the remaining items as a nonresponded
context set by assuming no click for this set. We augmented this set
of derived nonresponded sessions to the training set, and train the
model on the combined training data. The results from this
preliminary evaluation did not show significant performance
improvement. This is likely due to the fact that the surrogate
distribution is invalid. A detailed analysis with more realistic
data is the subject of future research.

\section{Related Work}
\label{sec:relatedwork}
Although a natural reflection of a user's preference is the process
of interaction with the recommender, to our knowledge, this
interaction data has not been exploited for learning recommender
models. Instead, research on recommender systems has focused almost
exclusively on learning the dyadic data. Particularly collaborative
filtering approaches only capture the user-item dyadic data with
explicit user actions while the context dyads are typically treated
missing values. For example, the rating-oriented models aim to
approximating the ratings that users assigned to items
\cite{SarwarWWW01,McLaughlinSIGIR04,SalMni08,AgaChe09,ChenKDD09,KorenIEEE09};
whereas the recently proposed ranking-oriented algorithms
\cite{WeiKarLe07,LiuYan08} attempt to recover the ordinal ranking
information derived from the ratings.

By exploiting past records of user-item dyadic responses for future
prediction based on either \emph{neighborhood based}
\cite{SarwarWWW01,McLaughlinSIGIR04,LiuYan08} or \emph{latent factor
based} methods
\cite{SalMni08,AgaChe09,ChenKDD09,KorenIEEE09,WeiKarLe07},
collaborative filtering approaches encode the collaboration effect
that similar users get similar preference on similar items. In this
paper, by leveraging the user-recommender interaction data, we show
that much better recommender performance can be obtained when a
local-competition effect underlying the user choice behaviors is
also encoded.

The multinomial logit model we present is derived based on the
random utility theory \cite{Luc59,Man75}. The model is
well-established and has been widely used for a long time in, e.g.\
psychology \cite{Luc59}, economics \cite{McF73,Man75} and marketing
science \cite{GuaLit08}. Particularly, \cite{GuaLit08} applied the
model to examine the brand choice of households on grocery data;
\cite{GenRec79} showed this model is theoretically and empirically
superior to the $\ell_2$ regression model. More recently, the
pioneering work of \cite{FleHos07} first applied the model to
characterize online choices in recommender system and investigated
how recommender systems impact sales diversity. Following these
steps, this work further employs the model to learn factorization
models for recommendation.

The Hinge formulation of CCF shows close connection to the pairwise
preference learning approaches widely used in Web search ranking
\cite{HerGraObe99}. Our model, however, differs from these content
filtering models \cite{HerGraObe99} in that instead of learning a
feature mapping as in \cite{HerGraObe99}, our model uses the
formulation for learning a multiplicative latent factor model.

\section{Summary and future research}
\label{sec:conclusion}
We  presented a framework for learning recommender by modeling user
choice behavior in the user-system interaction process. Instead of
modeling only the sparse binary events of user actions as in
traditional collaborative filtering, the proposed
collaborative-competitive filtering models take into account the
contexts in which user decisions are made. We presented two models
in this spirit, established efficient learning algorithms and
demonstrated the effectiveness of the proposed approaches with
extensive experiments on three large-scale real-world recommendation
data sets.

\smallskip\noindent There are several promising directions for future
research.

\paragraph{Attention budget and position bias.}
When deriving the CCF model, we admit an assumption that user
decides whether to take an offer solely based on the comparison of
utilities. This assumption, however, neglects a factor which might
be important in practice. In particular, a user might have budgeted
attention such that when making choices he only pays attention to a
few top-ranked items and totally disregard the others. This
\emph{position bias} is evident in both web search ranking and
recommendation. We plan to take this into consideration for building
choice models.

\paragraph{Recommender strategy and user behavior.}
A key feature of the current paper is that we assume the recommender
adopts a deterministic strategic policy when making recommendations.
In practice, a recommender could also adaptively react  to the
users' actions as well as its own considerations (e.g.\ inventory
constraints, promotion requirement of certain brands). We would like
to extend our analysis here to model the interactive process between
users and recommender.

\paragraph{Further empirical validation.} Due to data collection
constraints, some parts of the proposed models are not strictly
evaluated in the current paper. We plan to refine the mechanism for
data collection and conduct experiments for further experiments.

%{
%\bibliography{bib_ccf}
%}

\end{document}